\documentclass[conference]{IEEEtran}
\IEEEoverridecommandlockouts

\usepackage{algorithm}
\usepackage{array}
\usepackage[caption=false,font=normalsize,labelfont=sf,textfont=sf]{subfig}
\usepackage{stfloats}
\usepackage{url}
\usepackage{verbatim}
\usepackage{cite}
\usepackage{multirow}
\usepackage{booktabs}
\usepackage{threeparttable}

\usepackage{cite}
\usepackage{amsmath,amssymb,amsfonts}
\usepackage{algorithmic}
\usepackage{graphicx}
\usepackage{textcomp}
\usepackage{xcolor}
\def\BibTeX{{\rm B\kern-.05em{\sc i\kern-.025em b}\kern-.08em
    T\kern-.1667em\lower.7ex\hbox{E}\kern-.125emX}}

\begin{document}

\title{FAMNet: Integrating 2D and 3D Features for Micro-expression Recognition via Multi-task Learning and Hierarchical Attention
}

\author{
\IEEEauthorblockN{
Liangyu Fu\textsuperscript{1},
Xuecheng Wu\textsuperscript{2*}\thanks{*Corresponding author: wuxc3@stu.xjtu.edu.cn},
Danlei Huang\textsuperscript{2},
Xinyi Yin\textsuperscript{3}
}
\IEEEauthorblockA{
\textsuperscript{1}\textit{School of Software,
Northwestern Polytechnical University, Xi'an, Shaanxi, China
}}
\IEEEauthorblockA{
\textsuperscript{2}\textit{School of Computer Science
and Technology,
Xi’an Jiaotong University, Xi’an, Shaanxi, China
}}
\IEEEauthorblockA{
\textsuperscript{3}\textit{School of Cyber Science and Engineering,
Zhengzhou University, Zhengzhou, Henan, China
}}
\IEEEauthorblockA{
Emails: lyfu@mail.nwpu.edu.cn, wuxc3@stu.xjtu.edu.cn,\\
forsummer@stu.xjtu.edu.cn, yinxinyi@stu.zzu.edu.cn
}
}

\maketitle

\begin{abstract}
Micro-expressions recognition (MER) has essential application value in many fields, but the short duration and low intensity of micro-expressions (MEs) bring considerable challenges to MER. The current MER methods in deep learning mainly include three data loading methods: static images, dynamic image sequence, and a combination of the two streams. How to effectively extract MEs' fine-grained and spatiotemporal features has been difficult to solve. This paper proposes a new MER method based on multi-task learning and hierarchical attention, which fully extracts MEs' omni-directional features by merging 2D and 3D CNNs. The fusion model consists of a 2D CNN AMNet2D and a 3D CNN AMNet3D, with similar structures consisting of a shared backbone network Resnet18 and attention modules. During training, the model adopts different data loading methods to adapt to two specific networks respectively, jointly trains on the tasks of MER and facial action unit detection (FAUD), and adopts the parameter hard sharing for information association, which further improves the effect of the MER task, and the final fused model is called FAMNet. Extensive experimental results show that our proposed FAMNet significantly improves task performance. On the SAMM, CASME II and MMEW datasets, FAMNet achieves 83.75\% (UAR) and 84.03\% (UF1). Furthermore, on the challenging CAS(ME)$^3$ dataset, FAMNet achieves 51\% (UAR) and 43.42\% (UF1).
\end{abstract}


\section{Introduction}
After a long period of progress, human society has evolved vibrant ways of expressing emotions, among which facial expressions are more common and intuitive. According to the scale of facial expressions, we usually divide them into macro-expressions (MAs) and micro-expressions (MEs). MEs is a type of facial expression first discovered and named by Ekman\cite{ekman1969nonverbal} in 1969. Compared with MAs, it has the characteristics of short duration, imperceptibility, and downward movement intensity. Individuals produce MEs in an unconscious state and can reflect their actual psychological activities at the current moment. It can be applied to application scenarios such as clinical medicine, criminal investigation, and public security as an additional clue. In addition, in hot research directions such as human-computer interaction, social robots, and autonomous driving, the research on MEs also shows excellent application potential. With the wide application of neural networks in many research fields and the promotion of multidisciplinary integration, deep learning has become one of the essential technologies for micro-expression analysis.

Micro-expression recognition (MER) belongs to micro-expression analysis, which applies algorithms to identify the emotional categories (such as disgust, sadness, surprise, anger, fear, and happiness) of ME clips. Due to the particularity of MEs, MER based on deep learning is exceptionally challenging. In general, most MER methods require manual feature extraction, and they generally adopt the single-task learning mode, unable to extract subtle facial emotion changes more efficiently and sufficiently.

To this end, we innovatively propose an end-to-end model via multi-task learning and attention. It uses a data loading method combining static and dynamic images, and the results obtained from both of them after feature extraction by the neural network are fused in the decision layer. The multi-task in MER includes micro-expression recognition and facial action unit detection (FAUD). For the FAUD task, the facial action unit (AU) refers to the division of different regions of a human face from the perspective of human anatomy, each AU describes the apparent changes produced by a group of facial muscle movements, and its combination can express any facial expression. FAUD is a task that uses a machine learning algorithm to recognize the changes in facial action units accompanying a certain kind of expression. At the same time, we propose a hierarchical attention to extract fine-grained ME information more efficiently. The model is called FAMNet, where F, M, and A represent the fusion method, multi-task learning mode, and attention module, respectively. The main contributions of this paper are as follows:

\begin{itemize}
\item We propose a new end-to-end MER model FAMNet, which consists of two CNN branches, 2D and 3D, in which we deploy hierarchical attention and use multi-task learning to jointly train two specific tasks: MER and FAUD, to share the intermediate feature representation of MEs through hard sharing of parameters; the above approach enables the model to fully extract the comprehensive features of MEs and thus improve the model performance and generalization ability.
\item We innovatively construct a hierarchical attention module to focus on multiple intermediate feature representations in the backbone network, enabling the fine-grained spatial features of MEs to be extracted meticulously and efficiently.
\item Experiments on the MMEW, CAS(ME)$^3$, CASME II, and SAMM datasets show that our FAMNet exhibits superior performance compared to the existing state-of-the-art methods, with a maximum improvement of 3.6\% and 7.09\% under the UAR and UF1 evaluation metrics, respectively.
\end{itemize}

\section{Related Work}

\subsection{Traditional Micro-Expression Recognition.} 
Early MER approaches take feature extraction as the primary research direction, mainly utilizing traditional feature extraction methods, such as LBP, LBP-TOP and optical flow, to extract the feature representations and then deploying traditional machine learning algorithms for the sentimental classification. In 2009, Pfister et al.\cite{pfister2011recognising} first combined the Temporal Interpolation Model (TIM) algorithm and realized spontaneous MER based on LBP-TOP features. In 2015, Wang et al.\cite{wang2014lbp} proposed an improved LBP approach to reduce redundant information. Moreover, this model successfully achieves remarkable performance. In the same year, Liu et al.\cite{liu2015main} utilized an effective optical flow method to align micro-expression video sequences and deployed SVM to realize micro-expression classification. In addition, Liong et al.\cite{liong2014subtle} proposed a new feature extraction scheme and significantly improved the CASME II and SMIC datasets.

\subsection{Micro-expression recognition based on deep learning.} 
In recent years, various methods based on deep learning have made breakthroughs in the research field of MER. These methods mainly include three data loading methods: static images \cite{davison2018objective}, dynamic image sequences \cite{xia2019cross}, and a combination of both \cite{kumar2021micro}. Then these approaches deploy efficient neural networks to model the specific micro-expression dataset. For example, Zhou et al.\cite{zhou2019dual} constructed Dual-Inception neural network to realize MER across various databases. Van et al.\cite{van2019capsulenet} applied CapsuleNet for MER. Xie et al.\cite{xie2020assisted} used the graph neural network to improve model performance, combined with the facial action unit detection labels. Liong et al.\cite{8756567} proposed a lightweight three-dimensional model denoted STSTNet to extract deep emotional features. Xia et al.\cite{xia2019spatiotemporal} applied the RCN model to capture temporal and spatial information in the specific micro-expression sequences. Chen et al.\cite{chen2020spatiotemporal} utilized spatial attention and channel attention module to extract the spatio-temporal features of micro-expression sequences. In addition, to combat the sample imbalance problem, Peng et al.\cite{peng2018macro} fine-tuned the ResNet10 model on the macro-expression dataset and transferred it to the micro-expression dataset. Xie et al.\cite{xie2020assisted} proposed the AU-ICGAN model to enhance the feature representation of micro-expressions. Sun et al.\cite{sun2020dynamic} applied the knowledge distillation technology to transfer knowledge from the facial action unit. Nie et al.\cite{nie2021geme} introduced the Focal loss function to better deal with the unbalanced sample problem. In 2023, Chen et al.\cite{9676412} proposes a novel Block Division Convolutional Network (BDCNN) with implicit deep features augmentation. Based on these effective models and tricks, the accuracy of micro-expression recognition has been significantly improved more efficiently.

\begin{figure*}[t]
    \centering
\includegraphics[width=\textwidth]{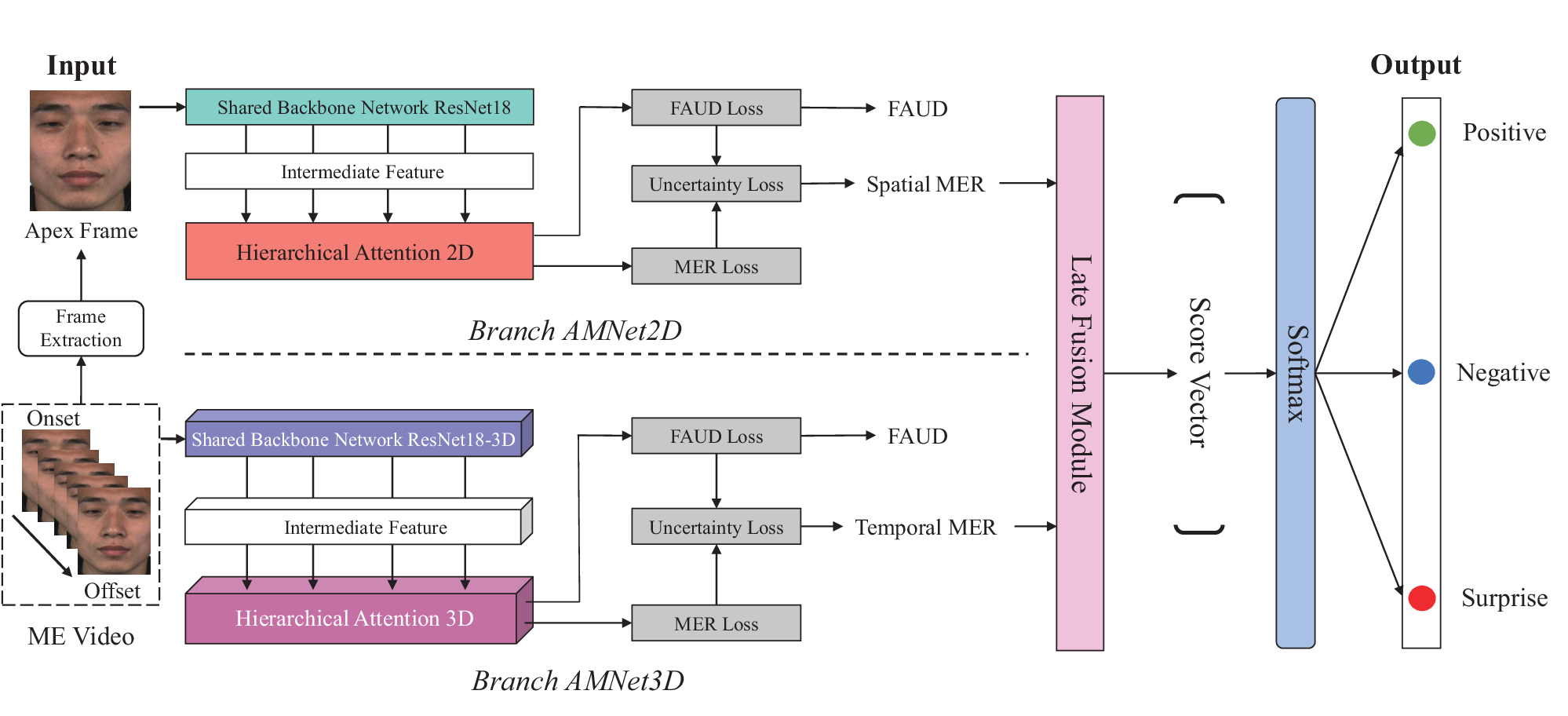}
	\caption{Overall framework of the proposed method. Specifically, the ``Input" consists of ``ME video" and ``Apex Frame" obtained by ``Frame Extraction". The two branches of FAMNet: AMNet2D and AMNet3D, are respectively input to extract the spatial and temporal features of MEs. The score vectors of the two branches are fused through the ``Late Fusion Module" to obtain the final ``Score Vector", and then the classification result is obtained through the ``Softmax" classifier. The detailed structure of network branches, attention modules and fusion will be elaborated in the following sections.}
\label{fig:overall}
\end{figure*}

\begin{figure*}[t]
    \centering
\includegraphics[width=\textwidth]{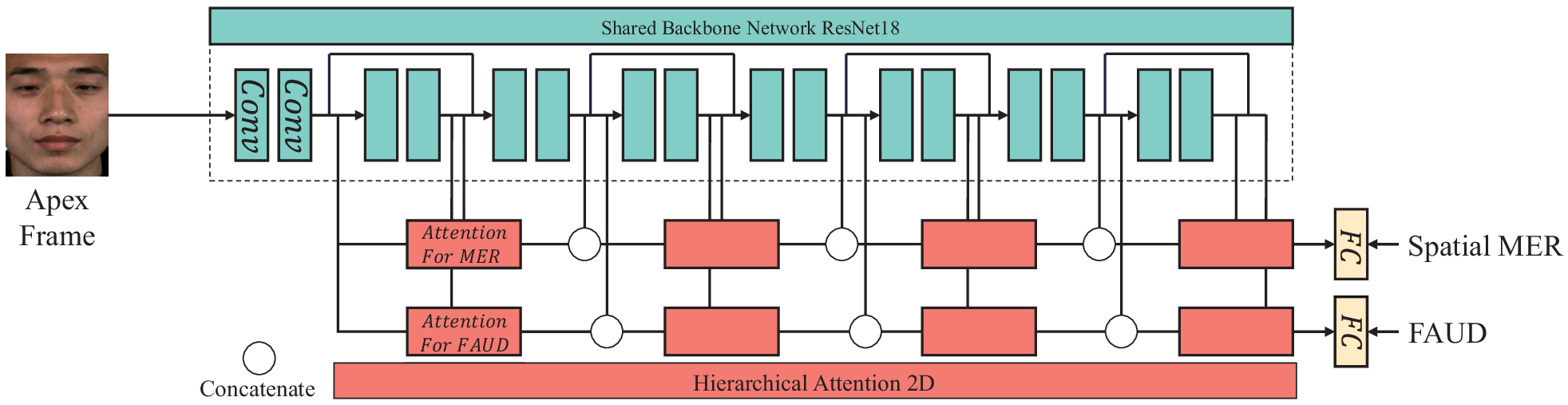}
	\caption{Network structure of AMNet2D. AMNet2D consists of the backbone network Resnet18 and hierarchical Attention, where ``Conv" represents the 2D convolutional layer, ``Attention For MER/FAUD" represents the attention module for two tasks, and ``FC" represents the fully connected layer. The hierarchical attention we proposed is applicable to two tasks, each of the four attention modules makes up a specific task attention module, and two specific task attention modules make up the hierarchical attention.}
\label{fig:AMNet2D}
\end{figure*}

\begin{figure*}[t]
    \centering
\includegraphics[width=\textwidth]{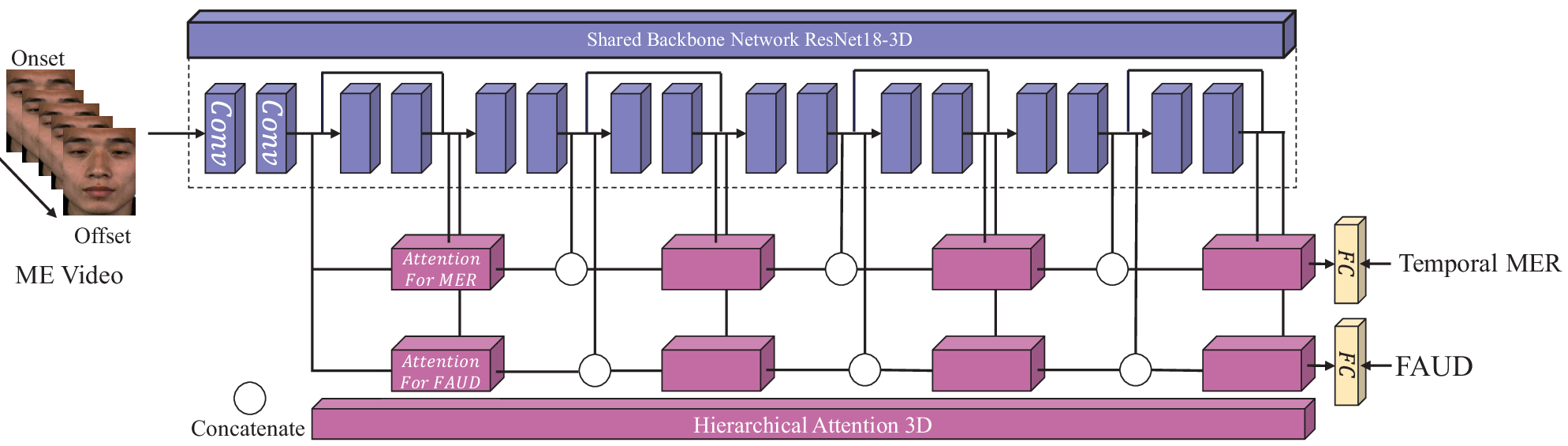}
	\caption{Network structure of AMNet3D. AMNet3D differs from AMNet2D in that all of its convolution operations (including in attention) have an additional dimension than AMNet2D, which is used to extract the temporal features of micro-expressions.}
\label{fig:AMNet3D}
\end{figure*}

\section{Method}
This paper proposes a new end-to-end fusion convolutional neural network model called FAMNet, based on multi-task learning and hierarchical attention. The overall structure of FAMNet is shown in Fig. \ref{fig:overall}. FAMNet consists of three parts: 2D CNN AMNet2D, 3D CNN AMNet3D, and late fusion module. There are differences in the way and dimension of data loading for the two networks. In addition, the attention module and multi-task learning mode are deployed in both. In the data loading module, for the 2D network, we use a single shared network to extract the deep feature representation of the Apex frames in the MEs sequence. In contrast, for the 3D network, we use a subject's MEs sequence loaded to the network. Next, we deploy hierarchical attention on the two network branches to pay detailed attention to the feature representations of multiple intermediate layers in the shared backbone network, extract fine-grained information, and finally classify after fusion at the FC layer. Furthermore, we use a multi-task learning mode to train MER and FAUD jointly. The following is divided into four parts to introduce the method: Network, Hierarchical Attention Modules, Fusion Module and Loss Functions.

\subsection{Branch Networks}
Convolutional neural networks have been widely used in image recognition, and their local receptive fields are usually used to extract local spatial features of images. The full name of ResNet is the residual network, this type of neural network is formed by stacking multiple residual modules, and a residual jump connection is designed between the residual modules. In our method, ResNet18 is selected as the shared backbone network, and the residual structure in ResNet18 is used to solve the problem of gradient disappearance and exploding problem. 

The two network branches in our proposed method use the 2D and 3D forms of ResNet18, respectively. 2D ResNet18 uses the default settings of He et al.\cite{he2016deep}, and 3D ResNet18 replaces all of the original convolutional blocks with 3D convolutional blocks. 


The specific structure of AMNet2D is shown in Fig.\ref{fig:AMNet2D}. Specifically, we take the Apex frame $X \in R^{c \times w \times h}$ that expresses the most emotional information in the MEs frame sequence as input, where $c,w,h$ denote the channel, width, and height of the images, respectively. The images are input to ResNet18 to get the output feature tensor $X_{4,2}^R \in R^{512 \times w/8 \times h/8}$. The general formulas are shown below:

\begin{equation}
X_{i,j}^R = Conv_{i,j}^R(X_{i-1,j+1}^R),
when j=1, i \in (1,2,3,4) \label{equ1}
\end{equation}
\begin{equation}
X_{i,j}^R = Conv_{i,j}^R(X_{i,j-1}^R),
when j=2, i \in (1,2,3,4) \label{equ2}
\end{equation}

Where $Conv_{i,j}^R$ denotes the convolutional transform block $Conv_{i,j}$ in ResNet18, and $X_{i,j}^R$ denotes the feature representation of the output of $Conv_{i,j}$, the intermediate layer of ResNet18. Specifically, $X_{0,2}^R = X$.

\subsubsection{AMNet3D}
In the 3D network, we use the MEs frame sequence(it can be broadly seen as a short micro-expression video) as the object of feature extraction to extract the temporal features of the MEs. The specific structure of AMNet3D is shown in Fig.\ref{fig:AMNet3D}. Specifically, unlike the AMNet2D above, we take the MEs video $V \in R^{c \times w \times h \times d}$ as input, where $c,w,h,d$ represent the input channel, width, height, and depth, respectively. The depth here is generally the length of the video, and then input the image into the 3D Resnet18 to obtain the output $V_{4,2}^R \in R^{512 \times w/8 \times h/8 \times d/8}$. 3D Resnet18 deploys a $3 \times 3 \times 3$ convolution kernel configuration. The overall formula is shown as follows:
\begin{equation}
V_{i,j}^R = Conv_{i,j}^R(V_{i-1,j+1}^R),~when~j=1, i \in (1,2,3,4),
\label{equ3}
\end{equation}
\begin{equation}
V_{i,j}^R = Conv_{i,j}^R(V_{i,j-1}^R),~when~j=2, i \in (1,2,3,4),
\label{equ4}
\end{equation}
where $Conv_{i,j}^R$ denotes the convolutional transform block $Conv_{i,j}$ in 3D ResNet18, and $V_{i,j}^R$ denotes the feature representation of the output of $Conv_{i,j}$, the intermediate layer of 3D ResNet18. Specifically, $V_{0,2}^R = V$.

\subsection{Hierarchical Attention}
The attention was widely used to solve problems in natural language processing in the early days and achieved remarkable results. In recent years, it has also been applied to computer vision. The primary purpose of attention in computer vision is to focus on different local areas in the image to improve the effect of feature extraction, just like in natural language processing, so that the model focuses on specific parts of the sentence instead of the entire sentence. It can focus on task-related regions while ignoring some background information irrelevant to the task. We deploy hierarchical attention modules in FAMNet to focus on fine-grained information related to MER. The specific structure of a single attention module is shown in Fig. \ref{fig:attention}. It uses a soft attention to calculate the attention matrix, and its specific working principle is:

\begin{figure}[t]
    \centering
\includegraphics[width=0.49\textwidth]{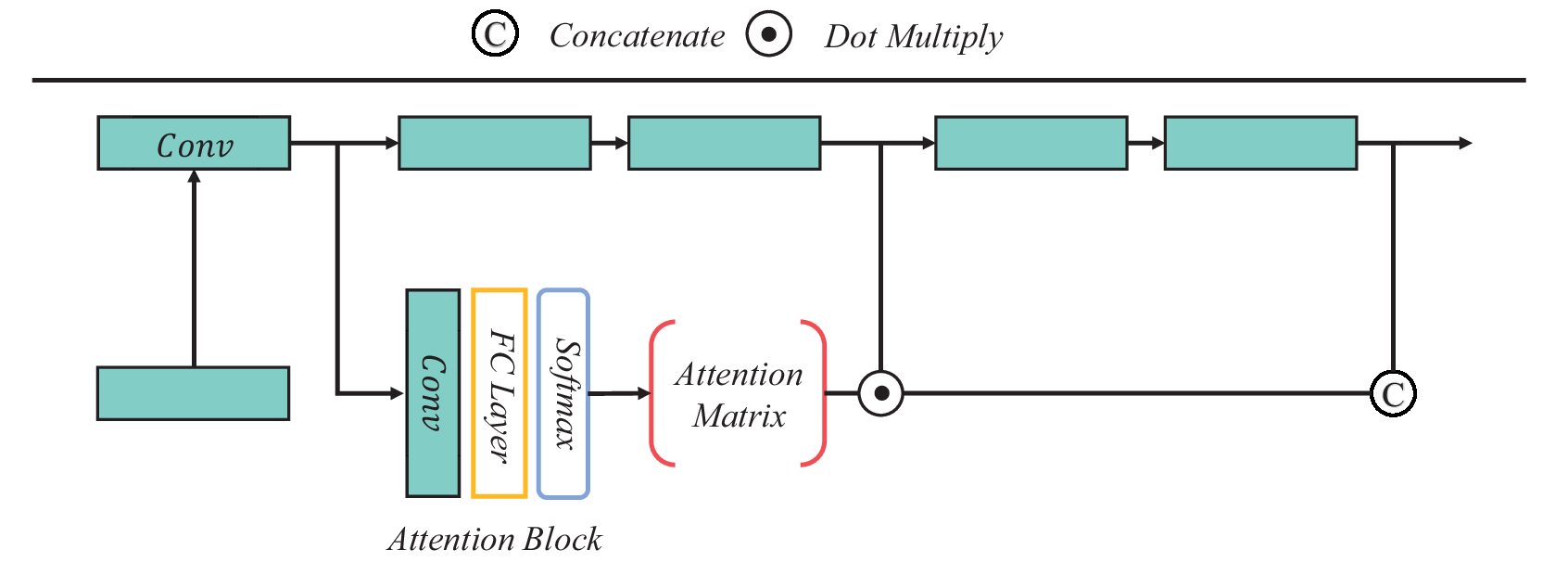}
	\caption{The specific structure of the attention module. Precisely, our proposed attention module consists of an attention block and an attention matrix; the attention block consists of a convolutional layer, a FC layer, and a layer of softmax activation functions; four attention modules in the above figure form a task-specific attention module, AMNet2D/3D contain two task-specific attention modules respectively, and they are collectively referred to as hierarchical attention.}
\label{fig:attention}
\end{figure}

For AMNet2D, set $X_{i,j}^R$ to denote the intermediate feature representation of the Apex frames passing through the shared network, where $i \in (1,2,3,4)$, $j \in (1,2)$. The formulas are shown below:

\begin{equation}
M_1 = softmax(Conv_1^A(X_{1,1}^R)),
\label{equ5}
\end{equation}
\begin{equation}
F_1 = X_{1,2}^R \odot M_1,
\label{equ6}
\end{equation}
\begin{equation}
M_k = softmax(Conv_k^A((F_{k-1},X_{k,1}^R)),
\label{equ7}
\end{equation}
\begin{equation}
F_k = X_{k,2}^R \odot M_k.
\label{equ8}
\end{equation}

For AMNet3D, set $V_{i,j}^R$ to denote the intermediate feature representation of MEs videos passing through the network, where $i \in (1,2,3,4)$, $j \in (1,2)$. The formulas are shown below:

\begin{equation}
M_1 = softmax(Conv_1^A(V_{1,1}^R)),
\label{equ9}
\end{equation}
\begin{equation}
F_1 = V_{1,2}^R \odot M_1,
\label{equ10}
\end{equation}
\begin{equation}
M_k = softmax(Conv_k^A((F_{k-1},V_{k,1}^R)),
\label{equ11}
\end{equation}
\begin{equation}
F_k = V_{k,2}^R \odot M_k,
\label{equ12}
\end{equation}
where Equation \ref{equ5}, \ref{equ9} and Equation \ref{equ6}, \ref{equ10} are the formulas for the first attention module in a single task, $Conv_1^A$ is the convolutional neural network in the first attention module, and $M_1$ is the attention matrix obtained from the first attention module. Equation \ref{equ7}, \ref{equ11} and Equation \ref{equ8}, \ref{equ12} are the formulas for the attention modules in a single task except for the first attention module. $Conv_k^A$ is the convolutional neural network in the $k$ th attention module, the input to the network is the channel dimensional concatenation of the matrix $F_{k-1}$ and the matrix $X_{k,1}^R$/$V_{k,1}^R$ and $M_k$ is the attention matrix computed by the $k$ th attention module, where $k \in (2,3,4)$. $Softmax(\bullet)$ denotes the softmax activation function and $\odot$ denotes the elemental multiplication of the matrix.

FAMNet has the same network structure for MER and FAUD. These attention modules are called hierarchical attention based on their layer-by-layer distribution structure in the neural network.

\subsection{Late Fusion Module}
\begin{figure}[t!]
\centering
\includegraphics[width=0.48\textwidth]{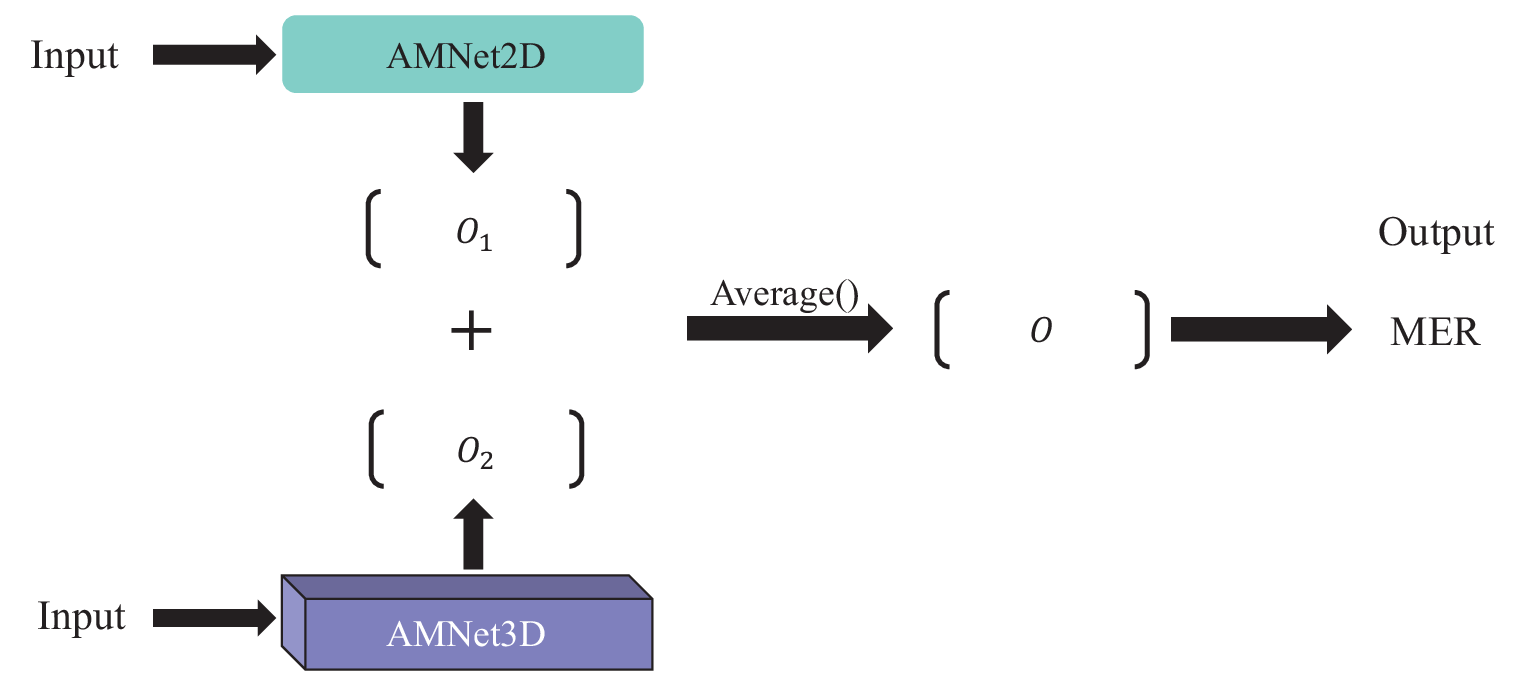}
\caption{The specific structure of the fusion module. Among them, ``$O_1$" and ``$O_2$" denote the prediction score matrices output from the FC layer of the MER task after feature extraction by AMNet2D and AMNet3D, respectively, and ``$O$" denotes the brand-new scores obtained after the matrix mean-taking operation. Finally, the classification results of MEs are obtained by the Softmax function.}
\label{fig:fusion}
\end{figure}

Our study is initially based on a 2D recognition model with multi-task learning. Apex frames of MEs sequences from the CASME II and SAMM datasets were put into the network for training, and the 2D model showed good performance. However, although the duration of MEs is short, it also possesses a complete process from occurrence to end, which does not appear suddenly at a specific moment, so our task should not be limited to Apex frames for MER. Inspired by the 3D model, we improved the original 2D model into a 3D model. A second data stream is added to the method to enable the model to extract the temporal features of MEs. In the fusion module, we fuse the prediction scores derived from the two branches in the FC layer to fully extract the spatiotemporal features of MEs to enhance the model's robustness. The fusion module is shown in Fig.\ref{fig:fusion}. Specifically, we take the prediction score matrix $O_1$ of the FC layer of AMNet2D and the prediction score vector $O_2$ of AMNet3D. Next, we calculate the average value of the two to obtain the output $O$, and finally the classification are derived; the specific formula is as follows:
\begin{equation}
 O=AVG(O_1, O_2),
\label{equ13}
\end{equation}
where $AVG()$ means averaging over the matrix elements.

\subsection{Loss function}
In the manner of multi-task learning, loss functions play a very important role in adjusting the training speed of multiple tasks to achieve the best learning results. In our proposed MHNet, the Uncertainty Loss Function is deployed to balance the loss values of two specific tasks.

Specifically, during the training process, when the loss function of a task has an upward trend, its corresponding parameter $\sigma$ will shrink to achieve an overall balance and optimize the specific task, where $\sigma_1$ and $\sigma_2$ are assigned to MER and FAUD, respectively. The goal is to find the optimal solution for these two parameters during the model convergence process.

\textbf{Loss Function for MER}: MER is a single-label classification task, and the loss function is calculated as follows:

\begin{equation}
\begin{split}
\centering
L_{ME}(\widehat{Y}_{ME},Y_{ME}) & = Y_{ME} \times (1-\widehat{Y}_{ME})^2 \\
& + 0.5 \times (1-Y_{ME}) \times \widehat{Y}_{ME}^2,
\label{equ14}
\end{split}
\end{equation}
where $\widehat{Y}_{ME}$ denotes the predict label and $Y_{ME}$ denotes the ground-truth label.

\textbf{Loss Function for FAUD}: The facial action unit detection (FAUD) is a multi-label classification task. And we thus utilize the BCE loss as a multi-label loss function with the following equations:

\begin{equation}
l_n = -w_n[y_nlog(\widehat{y}_n)+(1-y_n)log(1-\widehat{y}_n)],
\label{equ15}
\end{equation}

\begin{equation}
L_{AU}(\widehat{Y}_{AU},Y_{AU})=mean({l_1,L_2,...,l_n}),
\label{equ16}
\end{equation}
where $y_n$ and $\widehat{y}_n$ denote the ground-truth label and the prediction results of the $n$ th category, respectively. Besides, $w_n$ and $l_n$ denote the category weight and the loss of the $n$ th category in the samples,respectively. $mean(\bullet)$ refers to the average calculation.

\textbf{Uncertainty Loss Function \cite{kendall2018multi}}: Combining the above output $L_{ME}$ and $L_{AU}$, the total loss function $L_{Uncertainty Loss}$ can be calculated by the Uncertainty Loss function with the following formulas, \textit{i.e.},

\begin{equation}
\begin{split}
L_{Uncertainty Loss} & = \frac{1}{\sigma_1^2}L_{ME}(\widehat{Y}_{ME},Y_{ME}) \\
& + \frac{1}{\sigma_2^2}L_{AU}(\widehat{Y}_{AU},Y_{AU})+log\sigma_1+log\sigma_2,  \end{split}
\label{equ17}
\end{equation}
where $\sigma_1$ and $\sigma_2$ are the two specific parameters of the Uncertainty Loss Function. The purpose of the Uncertainty Loss Function is to find the optimal parameters, which are $\sigma_1$ and $\sigma_2$, respectively.

\section{Experiment}
\subsection{Datasets}
In order to verify the effectiveness of our proposed method, this paper conducts experiments based on the CASME II, SAMM, MMEW and CAS(ME)$^3$ datasets, respectively, and the basic information of the datasets is as follows. 


\textbf{CASME II \cite{yan2014casme}} is a comprehensive spontaneous MEs dataset containing 255 MEs samples collected from 26 Asian subjects with an average age of 22.59 years. The dataset is labeled with seven categories of emotions: happiness, disgust, surprise, depression, sadness, fear, and others.

\textbf{SAMM \cite{davison2016samm}} is a spontaneous dataset comprising 159 MEs samples and 32 subjects from 13 ethnic groups with an average age of 33.24 years. The dataset is labeled with eight categories of emotions: happiness, anger, surprise, contempt, disgust, fear, sadness, and others.

\textbf{MMEW \cite{ben2021video}} follows the same heuristic paradigm as CASME II, SAMM, all samples are carefully calibrated by experts with start, apex and offset frames, and FACS's AU annotations are used to describe facial muscle movement regions. Compared with the former two, the samples in MMEW have larger image resolution, add more emotion categories, and have a larger sample size.

\textbf{CAS(ME)$^3$ \cite{9774929}} provides about 80 hours of video, over 8 million frames, containing 1109 manually annotated MEs. Such a large sample size allowed efficient validation of the MER method while avoiding database bias. Inspired by experiments in psychology, it provides unprecedented depth information as an additional model. CAS(ME)$^3$ is the first to elicit MEs with high ecological validity using a simulated crime paradigm and physiological and speech signals, contributing to practical MEA. In addition, it provides 1,508 unlabeled videos with more than 4,000,000 frames, providing a data platform for the unsupervised MER method.

According to the method of Van et al.\cite{van2019capsulenet}, the experiments in this paper divide the emotional labels in the datasets into three categories: Positive, Negative, and Surprise. In the CAS(ME)$^3$ and CASME II datasets, we classify happiness as positive, disgust, sadness, fear as negative, and surprise as a surprise. In the SAMM and MMEW dataset, happiness is classified as positive. Sadness, contempt, anger, fear, and disgust were categorized as negative. Also, surprises are classified as Surprises.

\subsection{Implementation details}
Four datasets are first pre-processed in the following approaches:

\textbf{Data Pre-Processing}: All samples in the original four datasets contain useless background information, for which this paper deploys the Dlib to truncate the unnecessary background in the same way as in \cite{yan2014casme}. The same data augmentation approaches as in \cite{van2019capsulenet} are applied by resetting the images to 234 × 240 and then cropping them to 224 × 224 and finally applying transformations such as random rotation, random horizontal rotation, and random luminance. Also, to combat the small sample problem, this paper enriches the datasets by taking the Apex frames and their six neighboring frames from the micro-expression sequences as the available data samples.

\textbf{System Configuration}: In this paper, the Adam optimizer is applied to update the model parameters, and the optimizer uses the default parameter settings. All the proposed models are trained with a single NVIDIA Tesla V100 with 32GB memory and are entirely built with the PyTorch 1.10.1 neural network framework.

\begin{table*}[t]
\begin{center}
\caption{Performance comparisons of the different state-of-the-art MER models on the CASME II, SAMM, MMEW and CAS(ME)$^3$ datasets.}
\label{table:Performance of the different methods on the CASME II and SAMM datasets}
\centering
    \begin{tabular}{ccccccccc}
\toprule
\multirow{2}{*}{\textbf{Methods}} & \multicolumn{2}{c}{\textbf{CASME II}}      & \multicolumn{2}{c}{\textbf{SAMM}}    & \multicolumn{2}{c}{\textbf{CAS(ME)$^3$}}    & \multicolumn{2}{c}{\textbf{MMEW}}   \\
\cline{2-9}
                         & UAR             & UF1             & UAR             & UF1    & UAR             & UF1       &UAR &UF1   \\
\midrule
LBP-TOP \cite{zhao2007dynamic}                  & 0.7429          & 0.7026          & 0.4102          & 0.3954    & 0.2139          & 0.2178     &0.6361 &0.6423  \\ 

VGG11 \cite{simonyan2014very}                    & 0.5381          & 0.5315          & 0.4056          & 0.2871      & 0.2764          & 0.2697  &0.5923 &0.6025   \\

ResNet18 \cite{he2016deep}                & 0.5441          & 0.5367          & 0.4322          & 0.4821    & 0.2713          & 0.2654   &0.5971 &0.5901    \\

CapsuleNet \cite{van2019capsulenet}               & 0.7018          & 0.7068          & 0.5989          & 0.6209     & 0.2977          & 0.2352   &0.6834 &0.6762  \\

STSTNet \cite{8756567}               & \textbf{0.8382}          & \textbf{0.8686}          & 0.6588          & 0.6810     & 0.3761    & 0.3723   &0.8253 &0.8037   \\

EMR \cite{8756583}                       & 0.8293          & 0.8209          & 0.7754          & 0.7152     & 0.3656          & 0.3613  &0.8266 &\textbf{0.8149}    \\
\textbf{Ours}            & 0.8375 & 0.8403 & \textbf{0.8012} & \textbf{0.7861} & \textbf{0.5100} & \textbf{0.4342} &\textbf{0.8359} &0.8111 \\
\bottomrule
\end{tabular}
\end{center}
\end{table*}

\subsection{Evaluation Metrics}
Due to the lack of samples and unbalanced label distribution, we take $UF1$ and $UAR$ as metrics, \textit{i.e.},
\begin{align}
UF1 &= \frac{1}{C}\sum_{c}\frac{2TP_c}{2TP_c+FP_c+FN_c}, \\
UAR &= \frac{1}{C}\sum_{c}\frac{TP_c}{N_c},
\end{align}
where $TP$, $FP$, and $FN$ represent True Positive, False Positive, and False Negative, respectively. $C$ is the number of ME classes, and $c$ is the class of MEs.

\begin{figure*}[t]
    \centering
\includegraphics[width=0.8\textwidth]{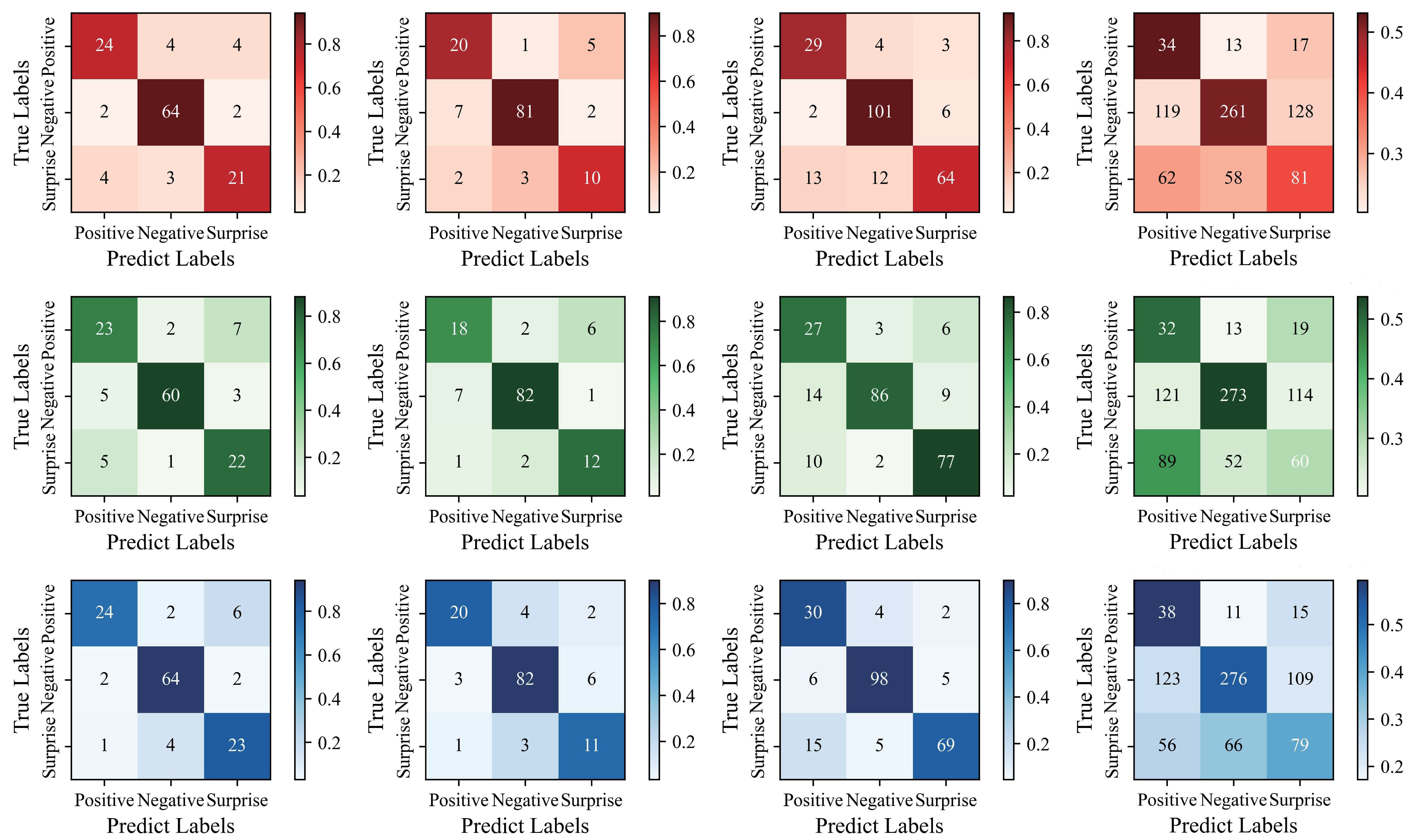}
	\caption{From top to bottom are the confusion matrices for AMNet2D, AMNet3D, and FAMNet; in each row, from left to right, are the confusion matrix results for the three methods on the CASME II, SAMM, MMEW and CAS(ME)$^3$ datasets, where ``True Labels" represent the actual labeled sentiment labels in the dataset and ``Predict Labels" represent the results predicted by FAMNet.}
\label{fig:confusion}
\end{figure*}

\subsection{Comparative Experiment}
To validate the performance of FAMNet proposed in this paper, we compared the experimental results of FAMNet with the other existing state-of-the-art methods mentioned above on the CAS(ME)$^3$, CASME II, MMEW and SAMM datasets. In all experiments, the learning rates were set to 1e-4, 4e-3, 1e-3 and 2e-3, and the exponential decay rates of the four datasets were 0.92, 0.9, 0.91 and 0.95, respectively. The same LOSO cross-validation method was used to train 100 epochs simultaneously. For the determination of the final performance of the model, we discard the highest value among all the UAR and UF1 data derived from the test set and select the next highest value as the determination of the final performance of FAMNet. The detailed performance comparison results are shown in Table \ref{table:Performance of the different methods on the CASME II and SAMM datasets}.

The experimental results in Table \ref{table:Performance of the different methods on the CASME II and SAMM datasets} show that FAMNet significantly improved performance over the existing state-of-the-art methods. To analyze the reason, existing methods merely deploy traditional manual features and extract micro-expression feature information from human faces using CNNs and single-task learning. In contrast, FAMNet focuses on fine-grained features of emotion-related MEs through hierarchical attention, applies multi-task learning to train MER and FAUD to explore information associations jointly, and fuses 2D and 3D networks to enable the model to extract spatiotemporal features of MEs, which effectively improves the generalization ability and robustness of the model using the above methods. In addition, STSTNet performs better on the CASME II dataset. We attribute this to the fact that in the method proposed by Liong et al., before using the neural network for feature extraction, they computed the optical flow guiding features using the onset and Apex frames, enabling the neural network to extract features in another dimension. This step plays an adequate effect on the CASME II compared to our method.

To better demonstrate the experimental results, the confusion matrix of the MER is shown in Fig.\ref{fig:confusion} based on the four datasets. It can be seen that the proposed FAMNet has high accuracy and generalizability: From the perspective of the results of the predicted categories, the highest accuracy rates of the positive, negative, and surprise categories are 0.833, 0.941, and 0.94, respectively; From the perspective of the four datasets, the negative category has the highest accuracy rate, which can reach 0.941, 0.901, 0.899 and 0.543, respectively.

\begin{table*}[t]
\centering
\caption{The Ablation Experiments for the Proposed FAMNet on the Task of MER, with the Single Shared Backbone Network ResNet18 as the Baseline Method for the Following Experiments}
\label{table:Ablation experiments for the micro-expression recognition}
\begin{tabular}{ccccccccc}  
\toprule
\multirow{2}{*}{\textbf{Methods}} & \multicolumn{2}{c}{\textbf{CASME II}} & \multicolumn{2}{c}{\textbf{SAMM}}  & \multicolumn{2}{c}{\textbf{CAS(ME)$^3$}}  &\multicolumn{2}{c}{\textbf{MMEW}}\\
\cline{2-9}
                         & UAR    & UF1                 & UAR    & UF1       & UAR    & UF1   & UAR    & UF1    \\
\midrule
Baseline                 & 0.5441 & 0.5367              & 0.4322 & 0.4821     & 0.2713 & 0.2054  & 0.5971 & 0.5831   \\ 
Single+HA(2D)                & 0.7649 & 0.7641              & 0.6900 & 0.648     & 0.4034 & 0.4073  & 0.6976 & 0.6947    \\ 
Dual+HA(2D)                & 0.8137 & 0.8196              & 0.7786 & 0.7593     & 0.4825 & 0.4148   & 0.8171 & 0.8029   \\ 
Single+HA(3D)                & 0.7861 & 0.7732              & 0.6831 & 0.6332     & 0.3983 & 0.3115   & 0.7912 & 0.8026   \\ 
Dual+HA(3D)                & 0.7956 & 0.7857              & 0.8011 & 0.7767     & 0.4452 & 0.3864   & 0.8013 & 0.7772   \\ 
FAMNet                  & \textbf{0.8375} & \textbf{0.8403}              & \textbf{0.8012} & \textbf{0.7861}    & \textbf{0.5100} & \textbf{0.4342}    & \textbf{0.8359} & \textbf{0.8111}   \\
\bottomrule
\end{tabular}
\end{table*}

\subsection{Ablation experiment}
Ablation experiment based on MER was performed to verify the proposed method's effectiveness in this paper. Its hardware, software environment and hyperparameter configuration are consistent with the comparison experiment. ResNet18 was set as the baseline model for the ablation experiments, using the original structure configuration except for the last layer. The statistics of the experimental results are shown in Table \ref{table:Ablation experiments for the micro-expression recognition}. Where ``Single+HA (2D)" denotes AMNet2D with a single-task learning method and hierarchical attention modules, ``Single+MA (3D)" denotes AMNet3D with a single-task learning method and hierarchical attention module, ``Dual+MA (2D)" denotes single AMNet2D, and ``Dual+MA (3D)" denotes single AMNet3D. ``Ours (Fusion)'' denotes the final approach, which is the FAMNet that fuses the two branches.

As seen from the Table \ref{table:Ablation experiments for the micro-expression recognition}, compared to the baseline method, all our proposed methods showed significant improvements. Specifically, by comparing the experimental results between "Single+HA" and "Dual+HA", we can conclude that the multi-task mode has significant advantages over the single-task, which plays a positive role in preventing overfitting and improving the generalization of the model, and the FAUD supports the optimization of MER. By comparing the results of 2D with 3D methods, we can find no significant difference in performance between them. The performance of the 2D model is generally better than 3D, which we attribute to using Apex frames as input, and the high-quality data allows the model to extract more comprehensive and detailed feature information of MEs. By comparing the final model with the results of the other five methods, we found another significant improvement in the performance of the model, which can demonstrate that by fusing a CNN focusing on the fine-grained and spatial features with another CNN focusing on the temporal features of MEs, the model can extract more comprehensive features, thus enhancing its robustness and generalizability.

\section{Conclusion}
In this paper, we propose a dual-stream CNN based on multi-task learning and attention mechanism for micro-expression recognition, design two CNN branches in 2D and 3D with late fusion, use the hierarchical attention mechanism to extract fine-grained MEs emotion features, and use multi-task learning to jointly train MER and FAUD to obtain a more robust model. Extensive experimental results on the CASME II, SAMM, MMEW and CAS(ME)$^3$ datasets show that our proposed FAMNet significantly improves robustness and generalization, outperforming the existing state-of-the-art methods.

\bibliographystyle{IEEEbib}
\bibliography{ijcnn2025references}

\end{document}